\documentclass{article}
\usepackage{ijcai23}

\usepackage{times}
\usepackage{soul}
\usepackage{url}
\usepackage[hidelinks]{hyperref}
\usepackage[utf8]{inputenc}
\usepackage[small]{caption}
\usepackage{graphicx}
\usepackage{amsmath}
\usepackage{amsthm}
\usepackage{booktabs}
\usepackage{algorithm}
\usepackage{multirow}
\usepackage{enumitem}

\usepackage{subfigure}
\usepackage[T1]{fontenc}

\usepackage{algorithmic}
\usepackage[switch]{lineno}

\title{A Hierarchical Approach to Population Training for Human-AI Collaboration}

\author{
Yi Loo$^1$
\and
Chen Gong$^1$\And
Malika Meghjani$^1$
\affiliations
$^1$Singapore University of Technology and Design (SUTD)\\
\emails
\{yi\_loo, chen\_gong\}@mymail.sutd.edu.sg,
malika\_meghjani@sutd.edu.sg,
}

\begin{document}

\maketitle

\begin{abstract}

    A major challenge for deep reinforcement learning (DRL) agents is to collaborate with novel partners that were not encountered by them during the training phase. This is specifically worsened by an increased variance in action responses when the DRL agents collaborate with human partners due to the lack of consistency in human behaviors. Recent work have shown that training a single agent as the best response to a diverse population of training partners significantly increases an agent's robustness to novel partners. We further enhance the population-based training approach by introducing a Hierarchical Reinforcement Learning (HRL) based method for Human-AI Collaboration. Our agent is able to learn multiple best-response policies as its low-level policy while at the same time, it learns a high-level policy that acts as a manager which allows the agent to dynamically switch between the low-level best-response policies based on its current partner. We demonstrate that our method is able to dynamically adapt to novel partners of different play styles and skill levels in the 2-player collaborative Overcooked game environment. We also conducted a human study in the same environment to test the effectiveness of our method when partnering with real human subjects. Code is available at \href{https://gitlab.com/marvl-hipt/hipt}{\textit{https://gitlab.com/marvl-hipt/hipt}}.       
    
\end{abstract}

\section{Introduction}

\begin{figure*}[t]
  \centering
  \includegraphics[width=0.18\linewidth,height=0.1\textheight]{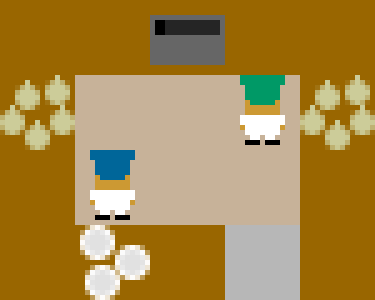}
  \includegraphics[width=0.24\linewidth,height=0.1\textheight]{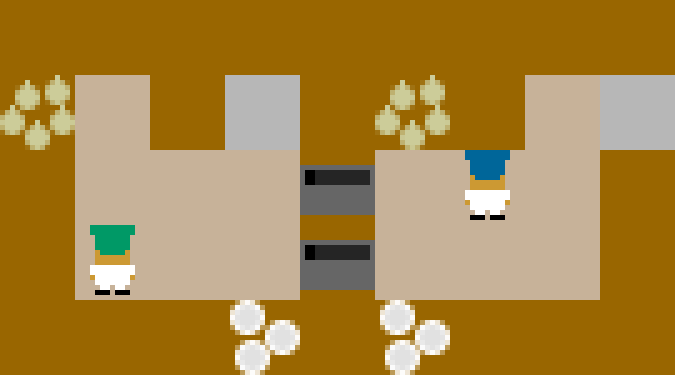}
  \includegraphics[width=0.16\linewidth,height=0.1\textheight]{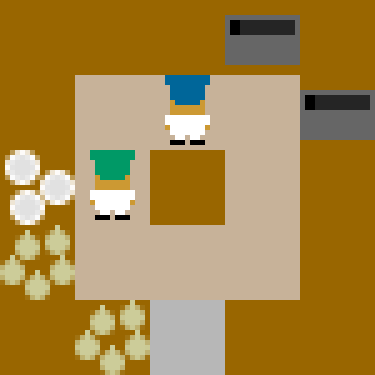}
  \includegraphics[width=0.16\linewidth,height=0.1\textheight]{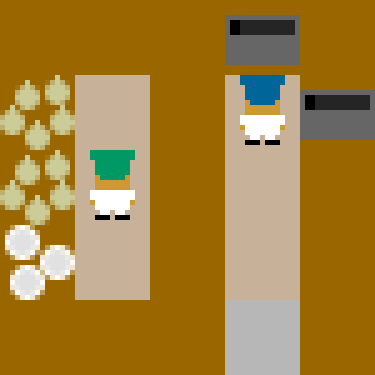}
  \includegraphics[width=0.24\linewidth,height=0.1\textheight]{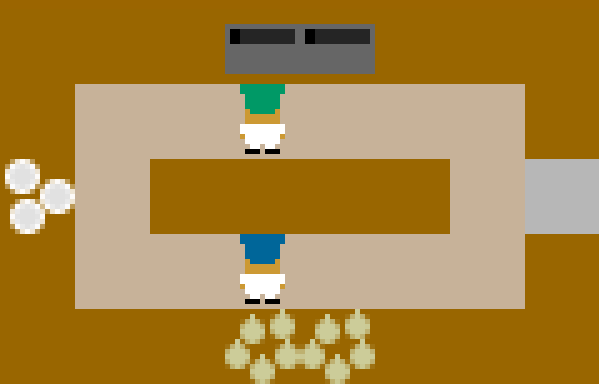}
  \caption{\textbf{The Five Overcooked layouts} From left to right: Cramped Room, Asymmetric Advantages, Coordination Ring, Forced Coordination and Counter Circuit. The Blue and Green hat chefs indicate the 2 different starting positions for each layout.}
  \label{fig:layout}
\end{figure*}

The idea of Intelligent agents that can robustly collaborate with humans has long been an aspirational goal in the field of Artificial Intelligence \cite{1363742} and Robotics \cite{alami2006toward}. To do so, agents would have to be able to dynamically adapt their behavior to novel unseen partners. The recent success of Deep Reinforcement Learning (DRL) methods has propelled them as a front-runner candidate in generating such robust collaborative agents. However the early successes of DRL mainly lie in the single-agent \cite{mnih2015human} or competitive domains \cite{silver2016mastering,berner2019dota}. In competitive settings, an optimal DRL agent would be able to outperform its opponent independently of the opponent's behavioral policy. However, DRL agents in cooperative/collaborative settings struggle to achieve good performance with novel partners unseen during training as DRL agents  develop cooperative conventions \cite{shih2020critical,hu2020other} that are only interpretable to its training partner. Multi-agent Reinforcement Learning (MARL) methods \cite{gronauer2022multi} do focus on cooperative domains but mainly deal with teams of agents that are trained jointly for a cooperative task  rather than having agents be robust to novel partners. In addition, many DRL methods struggle to generalize well to novel unseen situations \cite{kirk2021survey}. Recent work have shown that training a single RL agent as a best-response greatly increases agents' robustness to novel agent partners \cite{strouse2021collaborating,lupu2021trajectory,zhao2021maximum}. These methods however only consider partner diversity in terms of conventions/play-styles \cite{lupu2021trajectory,strouse2021collaborating}, (i.e. how different the agents' policies are from one another) or skill-level \cite{strouse2021collaborating}, (i.e. how capable the agent is at the collaborative task). We propose that truly robust agents would need to not collaborate well with optimal partners of different play styles but also sub-optimal partners of different skill levels

To that end, we introduce \textbf{\underline{Hi}erarchical \underline{P}opulation \underline{T}raining (HiPT)}, a population training method that leverages a hierarchical structure to learn multiple best response policies to collaborate with partners of different skill levels and play-styles. HiPT is inspired by ideas from Hierarchical Reinforcement Learning (HRL) \cite{pateria2021hierarchical}, a class of methods in DRL that impose a hierarchical structure on the agent. This hierarchical structure allows the agent to learn a task across multiple states and temporal abstractions. HRL methods typically learn high-level policies which learn multiple sub-goals of the task and lower-level policies that learn primitive actions which execute the said sub-goals. Our key insight is that multiple best-response policies correspond to sub-goals in the HRL framework and could be learned as low-level policies. The high-level policy meanwhile acts as a manager policy that allows the agents to dynamically switch between low-level, best-response policies to adapt to new partners. HiPT has a two-level hierarchical structure that can be trained with the popular PPO algorithm \cite{schulman2017proximal} while considering a diverse set of N population partners of varying play styles and skill levels.


We validate HiPT on the two-player cooperative game environment Overcooked \cite{carroll2019utility}.  We test our method on a diverse disjoint test set of Self-Play agent partners and  against agents trained to mimic human game-play using behavior cloning \cite{carroll2019utility}. We then conducted a large-scale human trial to test our method against real human partners. We show that  HiPT outperforms agents of other population-based methods \cite{strouse2021collaborating} and agents that are trained to adapt to human behavior \cite{carroll2019utility} in overall cooperative capability when partnering with both  trained agents and human partners.


We summarize the novel contributions of this work as follows:
\begin{enumerate}[label=(\alph*)]
    \item We propose HiPT, an HRL-based method for Human-AI Collaboration that is trained on a population of  diverse partners of different play styles and skill levels. (Section 4.1)

    \item We propose adding influence rewards to train the high-level policy to encourage the sub-policies to be more distinct. (Section 4.3)

    \item We show that HiPT outperforms similar Population-based training methods when partnering with trained agent partners and real human partners. In addition, HiPT is also shown to be preferred as a partner by human partners over FCP \cite{strouse2021collaborating} and $PPO_{BC}$ \cite{carroll2019utility} (Section 5)

\end{enumerate}


 
\section{Related Work}

There has been a growing number of works addressing the problem of \textbf{Human-AI Collaboration}. \cite{carroll2019utility} was the recent pioneering work that introduced the Overcooked AI environment and proposed to train PPO agents partnered with agents trained to mimic human gameplay. Though such a model is shown to be more robust to partnering with humans, it requires the collection of human data which could be quite expensive. Strouse et al. \shortcite{strouse2021collaborating} proposed a way to circumvent the process of collecting human data by training a single agent as the best response to a population of diverse partners of varying skill levels by using past checkpoints of the partners. Lupu et al. \shortcite{lupu2021trajectory} and Zhao et al. \shortcite{zhao2021maximum} also propose using the population training method to increase the robustness of agents in collaborative settings but focus on training partner populations that are diverse in terms of play style by either adding an additional loss term \cite{lupu2021trajectory} or modifying the reward function \cite{zhao2021maximum}. Our method also uses the population training approach but extends the definition of diversity to include both skill level and play style by adopting techniques from both branches of population-based methods. We also design an agent with a specialized hierarchical structure that can learn multiple best response policies to the partner populations.


A separate but related line of research considers the problem of \textbf{Zero-shot coordination (ZSC)} \cite{hu2020other,shih2020critical,hu2021off,lucas2022any}. Introduced by Hu et al. \shortcite{hu2020other}, ZSC the framework requires partner agents to be trained independently using the same method. The agents are then paired together and evaluated in cross-play (XP). This class of methods considers a narrower definition of a robust collaborative agent as it does not extend to agents trained with different methods or even human partners but is evaluated in more complex collaborative environments such as Hanabi \cite{bard2020hanabi}. Though more recent works have considered partners of different methods \cite{lucas2022any}. Our method in contrast solves the problem Human-AI collaboration and evaluates directly partnering with human partners.

 One other highly related area to our work is the field of \textbf{Hierarchical Reinforcement Learning (HRL)} \cite{BerliacHierachialRL2019,pateria2021hierarchical}. Sutton 
 et al. \shortcite{DBLP:journals/ai/SuttonPS99} first introduced the idea of learning a task at multiple levels of temporal abstraction in reinforcement learning. Since then, the field has evolved to include a wide range of methods such as the options framework \cite{stolle2002learning,bacon2017option} and feudal learning \cite{vezhnevets2017feudal}. Though HRL has mostly been adapted to the domain of single-agent environments, recent years have seen HRL methods being adapted to MARL setting, where teams of agents cooperate on a joint task \cite{yang2020hierarchical,wang2020rode}. To the best of our knowledge, our method is the first to apply HRL to the domain of Human-AI Collaboration or cooperation with unseen novel agents. Our method HiPT is most inspired by the HiPPO \cite{li2019sub} algorithm, which also learns a bi-level policy and is trained using the PPO objective. Our method however is designed for multi-agent cooperative settings with novel partners whereas the original HiPPO work is only evaluated in a single-agent setting. To that end, we reformulate the sub-policies as best response policies rather than individual skills. Furthermore,  we added an influence reward component to the high-level policy during training to encourage more distinct sub-policies for the low-level policy.

\section{Preliminaries}
\textbf{Multi-agent MDP} The problem of Human-AI collaboration can be formulated as a Multi-agent Markov Decision Process. A Multi-Agent MDP is defined by the tuple $\langle S, A, I, T, R\rangle$ where  $I$ is the set of agents. In this work, we restrict the number of agents to two as we only consider 2-player collaborative tasks. $S$ is the set of all possible states, and $s_t \in S$ is the state at time $t$. $A$ is the set of all possible actions and $a^i_t \in S$ is the action taken by agent $i$ at time $t$. The actions of all agents can be combined to form the joint action $\boldsymbol{a_t} = \langle a^1_t,...,a^I_t\rangle$ at time $t$. The transition function $T(s_{t+1} |s_t, \boldsymbol{a_t})$ determines the transition probability from state $s_t$ to $s_{t+1}$ given the joint action $\boldsymbol{a_t}$. Finally the reward function $R(s_t, \boldsymbol{a_t}, s_{t+1})$ is the reward that the agents receive after every state transition. The overall RL objective is for agents to maximize their discounted expected future reward $R^i = \sum_{t=0}^{\infty}\gamma^i r_t^i$ where $\gamma^i$ is each agents discount factor.

\medskip

\noindent \textbf{Hierarchical Proximal Policy Optimization (HiPPO)} \cite{li2019sub} is introduced as an augmentation to the Proximal Policy Optimization \cite{schulman2017proximal} algorithm by introducing a two-level hierarchical structure to the agent. The agent consists of a high level manager $\pi_{\theta_h}(z|s)$ and low-level sub-policy $\pi_{\theta_l}(a|z,s)$. The manager outputs a sub-policy command, $z$ which is the prior for the sub-policy network. The prior $z$ is kept constant for $p$ number of steps which is randomly sampled from a fixed range. Both the
high-level manager and the low-level sub-policy are trained using the PPO objective function.


\section{Method}
\begin{figure}[t!]
  \centering
  \includegraphics[width=0.8\linewidth]{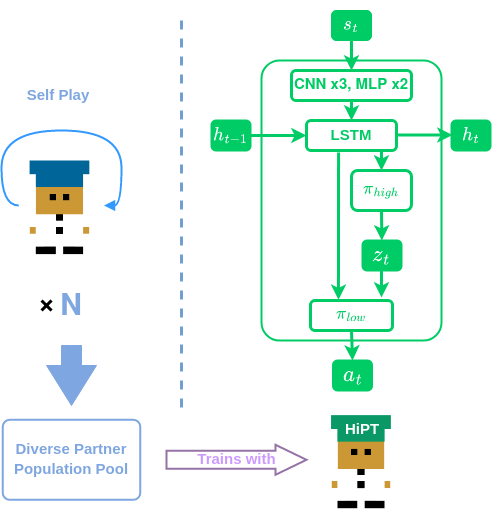}
  \caption{Overview of HiPT. We first construct a diverse partner population by training agents in self-play. We then train a HiPT agent to learn multiple best response policies from the partner population through the use of the agent's hierarchical structure.}
  \label{fig:overview}
\end{figure}

In this section, we describe the Hierarchical Population Training method in detail. We first describe the procedure of training a diverse partner population. We then describe the training procedure of HiPT agent itself. An overview of HiPT can be seen in Figure \ref{fig:overview}.

\begin{figure*}[ht]
  \centering
  \subfigure[Cramped Room]{\includegraphics[width=0.25\linewidth]{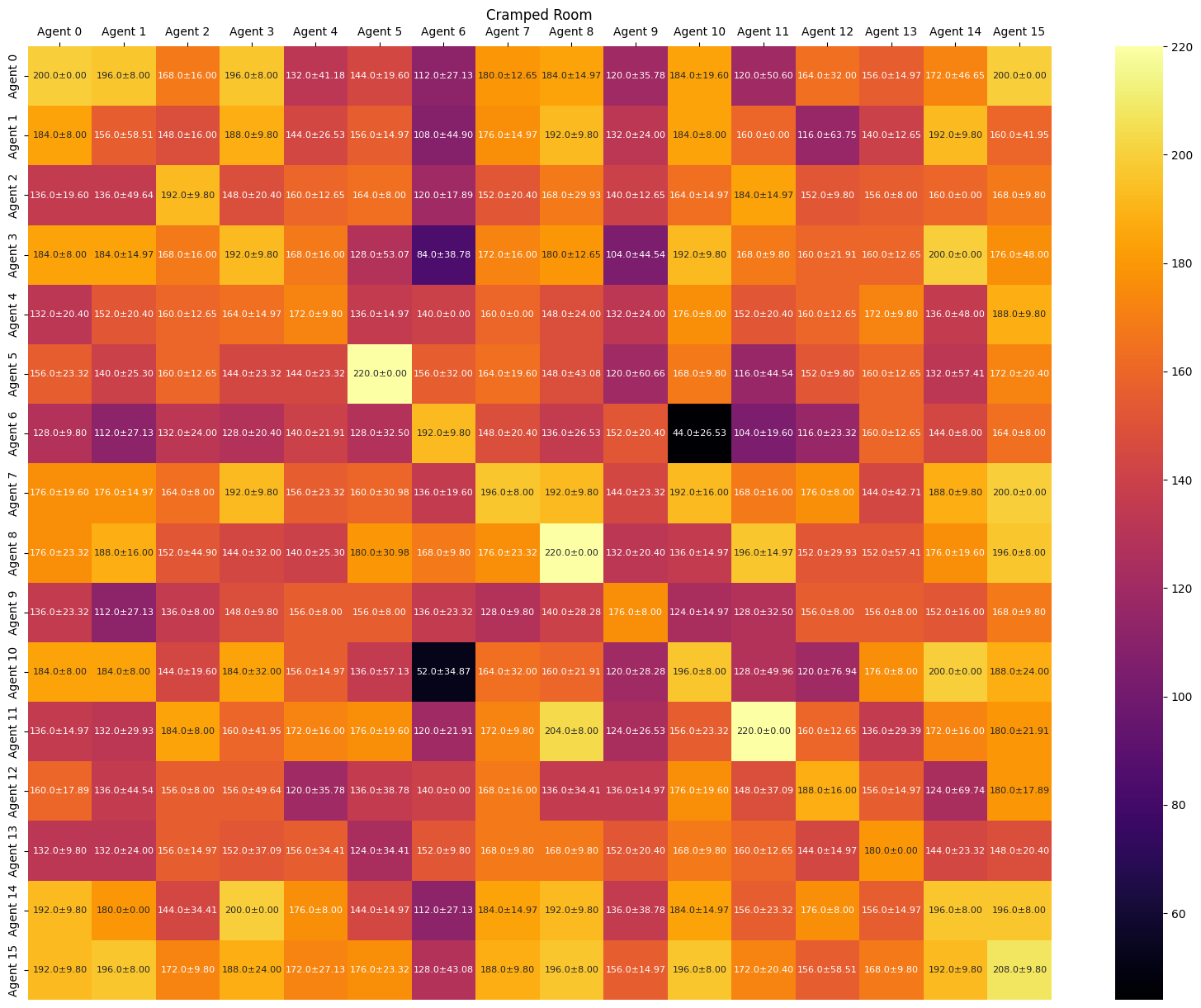}}
  \subfigure[Asymmetric Advantages]{\includegraphics[width=0.25\linewidth]{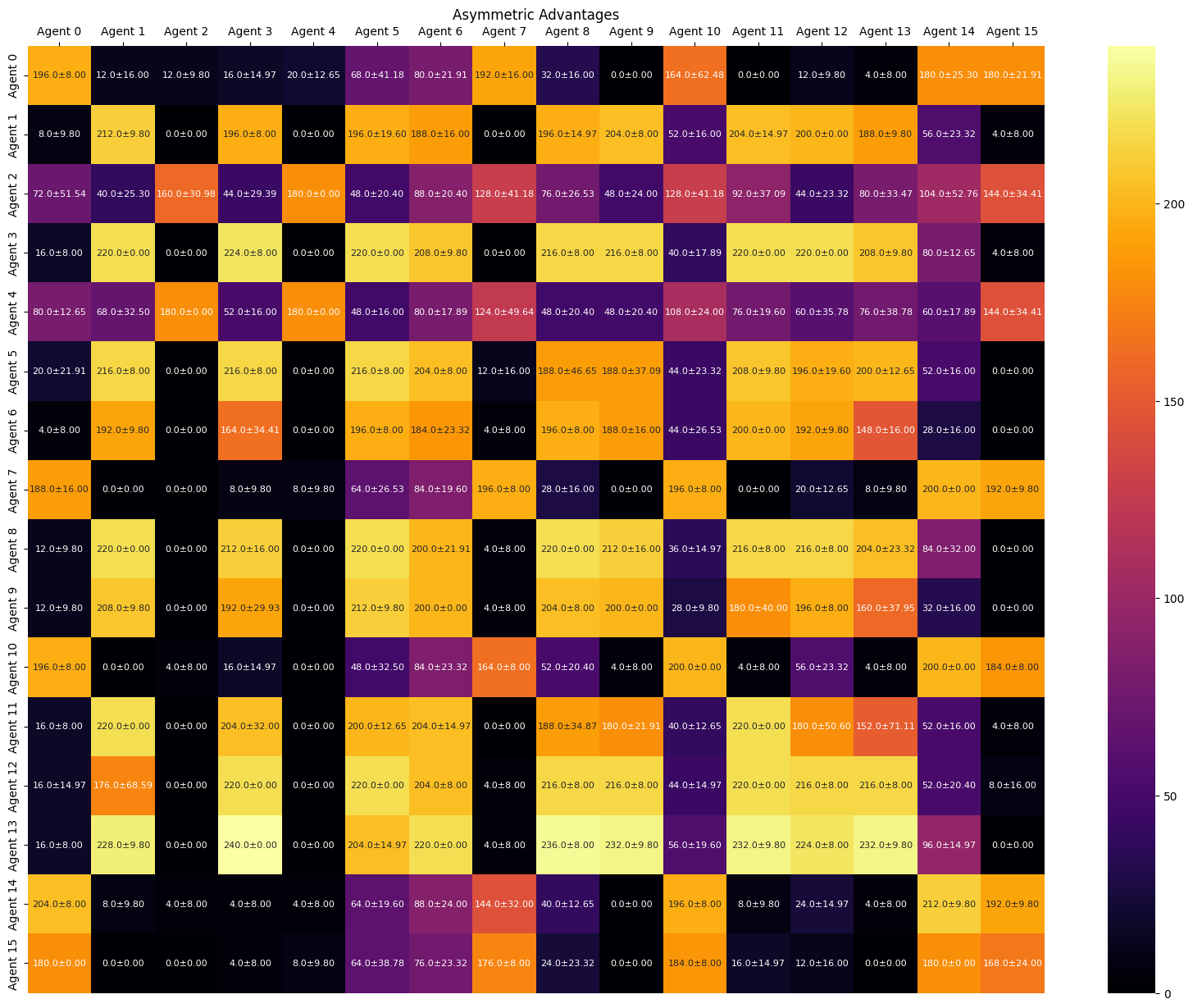}}
  \subfigure[Coordination Ring]{\includegraphics[width=0.25\linewidth]{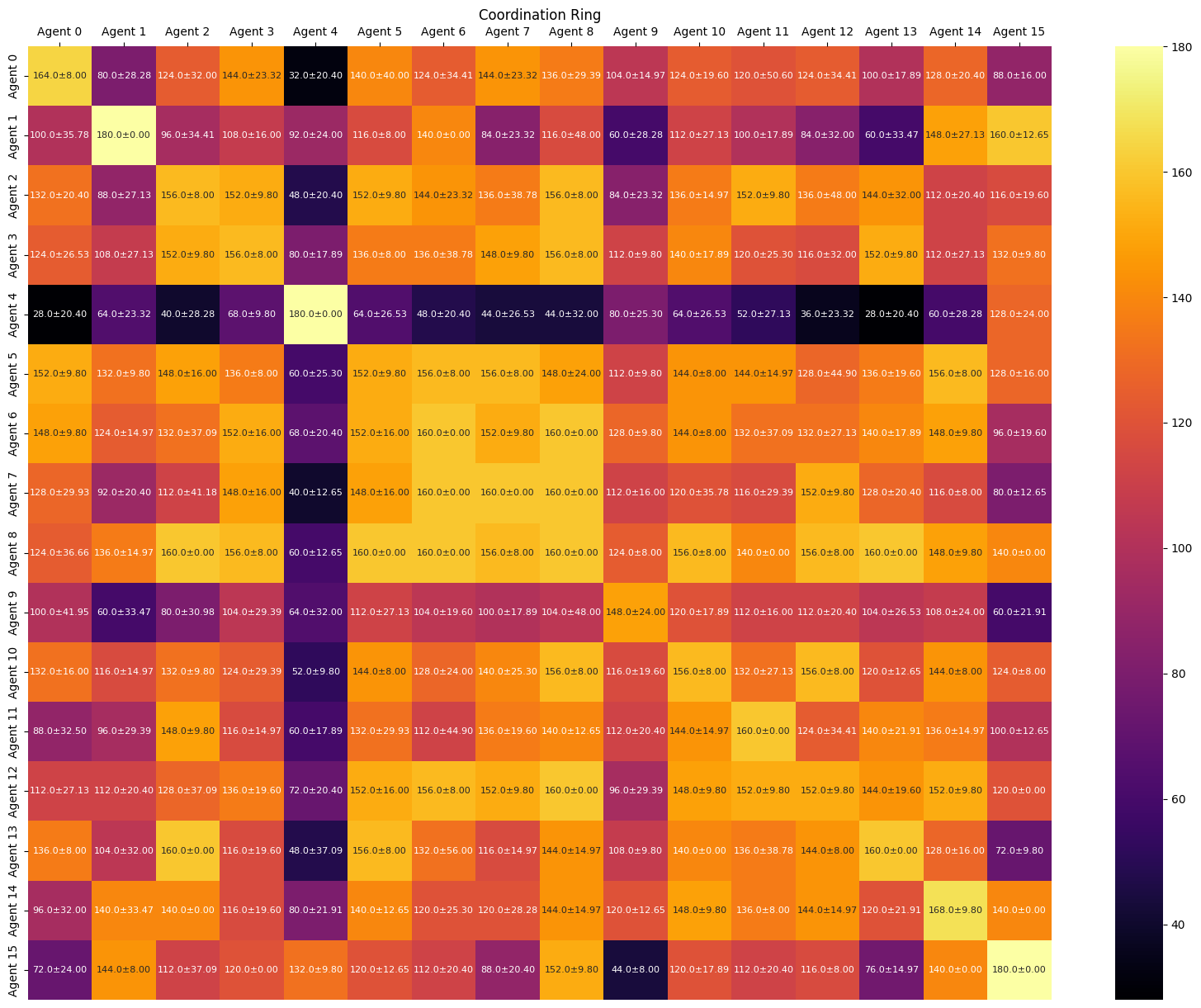}}
  \subfigure[Forced Coordination]{\includegraphics[width=0.25\linewidth]{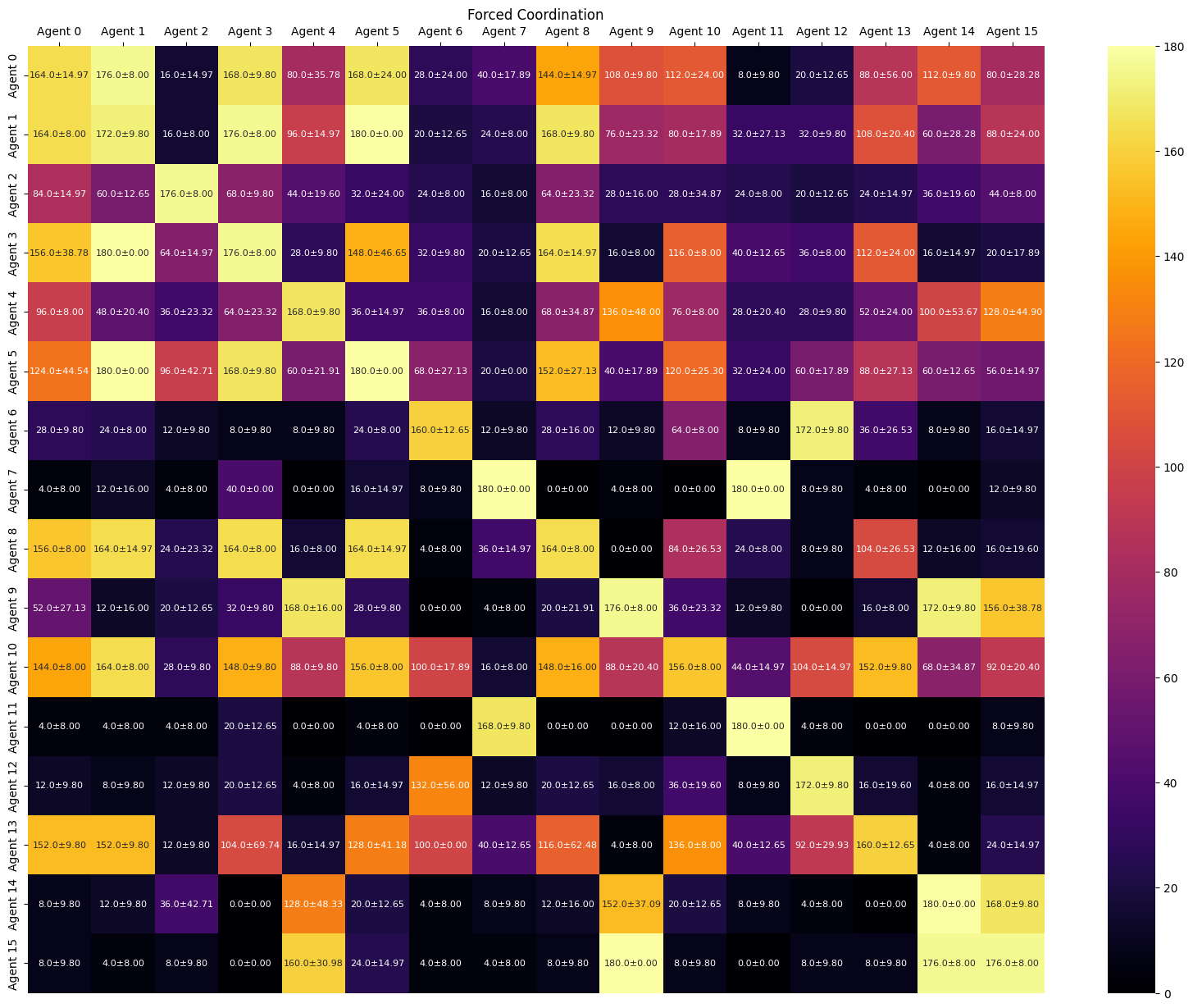}}
  \subfigure[Counter Circuit]{\includegraphics[width=0.25\linewidth]{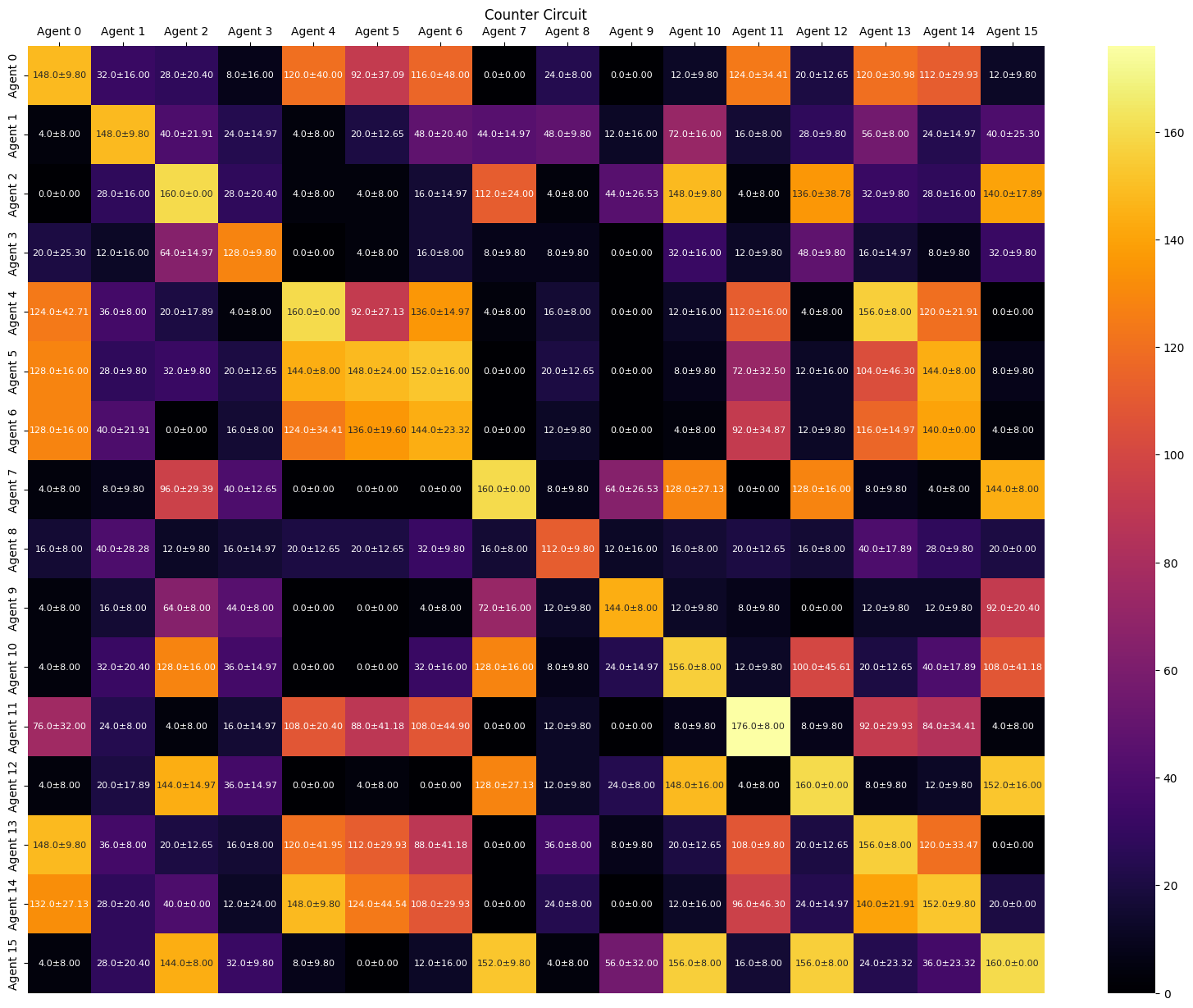}}
  \caption{\textbf{Heat map showing the average pairwise return of a Diverse Partner Population in Overcooked} across 5 separate layouts. Rewards are calculated over 5 episodes of length $T=400$ Lighter colored squares represent pairs that achieve higher reward, which indicates that the two agents belong to the same class of play-styles, while darker color square pairs that achieve comparatively low reward, indicating that they are incompatible in terms of play-style. The heatmap shows the population training is able to generate agents with groups of distinct play styles in the Asymmetric Advantages, Forced Coordination, and Counter Circuit layouts, while the classes in Cramped Room and Coordination Ring are less distinct possibly due to the simplicity of the 2 layouts that do not require much variation in play styles. (See Figure \ref{fig:layout} for reference to each layout)}.
  \label{fig:spheatmap}
  \vspace{-10pt}
\end{figure*}


\subsection{Diverse Partner Population Training}
For the training of the partner population, we train a pool of randomly seeded agents using PPO in SP. We consider the diversity of partners across two dimensions: 
\begin{itemize}
 \item \textbf{Play Styles} Formally, we define play style similar to how it is described in \cite{lupu2021trajectory} referred to as the Best Responses (BR) classes hypothesis, two agents, represented by policies $\pi_1$ and $\pi_2$ are said to belong to the same class/share the same play style if the average return of pairing $\pi_1$ and $\pi_2$, $J(\pi_1,\pi_2)$ is similar to that of the average self-play return of $\pi_1$ and $\pi_2$: $J_{SP}(\pi_1)$ and $J_{SP}(\pi_2)$, i.e. $J(\pi_1,\pi_2) \approx J_{SP}(\pi_1) \approx J_{SP}(\pi_2)$. 

\item \textbf{Skill Level} We define an agent's skill level simply by its average self-play return $J_{SP}(\pi)$, i.e. a highly skilled agent will obtain a high return in self-play and vice versa.
\end{itemize}

To ensure that agents in the population will have a diverse set of \textbf{play styles}, we adopt the technique proposed \cite{lupu2021trajectory} by adding a negative Jensen Shannon Divergence (JSD) term to the population training objective. The JSD term is calculated by:

\begin{align}
    JSD = \mathcal{H}(\hat{\pi}) - \frac{1}{N}\sum_{n=1}^{N} \mathcal{H}(\pi_n)
\end{align}

 \noindent where $\mathcal{H}(\pi)$ denotes the entropy of the policy and $\hat{\pi}$ is the average policy of the agent population. Intuitively this term encourages different play styles as the more dissimilar the policies are from one another, the higher the value of the JSD term. 

To ensure that the partner population contains partners of different \textbf{skill-level}, we follow the procedure proposed by Strouse et.al. \shortcite{strouse2021collaborating} where past agents corresponding to checkpoints agents are included to represent less skilled partners. More specifically, for each agent trained using self-play, we add three versions of the agent to the partner population: the fully trained agent, an agent from a past checkpoint that achieves around half the average return as the fully trained agent, and a randomly initialized agent.

We showcase the resultant diverse partner population by calculating the pairwise crossplay average return of a trained partner population of size $16$. We then plotted the returns in the form of a matrix heatmap and show them in Figure \ref{fig:spheatmap}.

\subsection{Hierarchical Population Training}


HiPT learns a high-level policy $\pi_{\theta_{high}}(z|s)$  and low-level policy  $\pi_{\theta_{low}}(a | z, s)$. $\pi_{\theta_{high}}(z|s)$ takes in the environment state, $s$ and outputs the sub-policy prior, $z$ while $\pi_{\theta_{low}}(a | z, s)$ takes in both $s$ and $z$ and outputs the primitive action $a$. Each sub-policy $z$ is executed for a random $p$ number of steps which are sampled uniformly from a lower and upper bound $(p_{min},p_{max})$.

In the context of HiPT, the low-level policy represents the best response policies to the partner population, conditioned by the high-level prior $z$. The high-level policy can then be thought of as a manager policy that chooses the most suitable BR policy as a response to the current partner's behavior. To further ensure that each learned BR sub-policies results in a distinct shift in behavior in the low-level policy we add an additional influence reward to the high-level policy which we detail in the following sub-section.

During the policy rollout stage of training, we uniformly sample a partner policy, $\pi_{p_n}$ from the diverse partner population. We then roll out an entire trajectory episode with $\pi_{p_n}$ as the partner.

\subsection{Influence Reward for High-level policy }


As the high-level manager policy is responsible for picking the most suitable BR sub-policy, We would like to ensure each sub-policy prior, $z_j$ results in a distinct change in the low-level policy. In other words, we would like to encourage each $z_j$ to have high influence over the behavior of $\pi_{\theta_{low}}(a|s,z)$.

To that end, inspired by \cite{jaques2019social}  during training we award influence reward, $r^{inf}$ to the high-level policy in addition to the base reward received from the environment. The influence reward at each time step is calculated by:

\begin{equation}
    \begin{aligned}
    r^{inf}_{t}& = D_{KL} [\pi_{\theta_{low}}(a_t | z_t, s_t) || \sum_{z \in Z} \pi_{\theta_{low}}(a_t | z, s_t)\pi_{\theta_{high}}(z|s_t)] \\
               & = D_{KL} [\pi_{\theta_{low}}(a_t | z_t, s_t) || \pi_{\theta_{low}}(a_t |  s_t)]
    \end{aligned}
\end{equation}

\noindent where $D_{KL}[\cdot||\cdot]$ is KL-divergence between two distributions. In our case, we are calculating the divergence between low-level policy given prior $z_t$ and marginal policy low-level policy $\pi_{\theta_{low}}(a_t |  s_t)$. 

The influence reward models how much a given sub-policy prior, $z$ influences the low-level policy, $\pi_{\theta_{low}}(a|s,z)$. The high-level policy will receive a higher influence reward if the given sub-policy prior $z_t$ results in a significantly different low-level policy at time $t$ as compared to a low-level policy that is not given prior $z_t$. The marginal policy $\pi_{\theta_{low}}(a_t |  s_t)$ can be computed by taking the expectation of the product between the low and high-level policies over the set of priors. Both distributions can be inferred directly at every time step given a discrete set of $z$ values. The full reward for the high-level policy is calculated by:

\begin{equation}
    \begin{aligned}
        r^{h}_{t} = \frac{1}{p}\sum_{t' = t}^{t+p} (\alpha r_{t'} + \gamma r^{inf}_{t'})
    \end{aligned}
\end{equation}

\noindent where p is the random step duration of the previous sub-policy. $\alpha$ and $\gamma$ are the coefficients for the environment and influence reward respectively. As the high-level policy outputs $z$ at a random number of steps, we average out the high-level reward over p to normalize the value range of $r^{h}_{t}$.  

We further validate the effect of adding the influence reward in an ablation study detailed in Section 5.3. The full rollout algorithm for HiPT is detailed in Algorithm \ref{alg:algorithm}. Note that the high-level policies and low-level policies operate at different temporal frequencies, resulting in high level trajectories that are shorter than the episode length, $T$.

\begin{algorithm}[tb]
    \caption{HiPT Rollout}
    \label{alg:algorithm}
    \textbf{Input}: Set of $N$ agent partners $\{\pi_{p_1},...,\pi_{p_N}\}$, high level policy $\pi_{\theta_{high}}(z|s)$, low-level policy $\pi_{\theta_{low}}(a|s,z)$ \\
    \textbf{Parameter}: Execution bounds for low-level policy $[p_{min},p_{max}]$, Influence reward coefficient $\gamma$, Environment reward coefficient  $\alpha$
    \begin{algorithmic}[1] 
        \STATE Sample random partner $\pi_{p_n} \sim \{\pi_{p_1},...,\pi_{p_N}\}$
        \FOR {$t=0$ \TO $T$}
        \STATE Sample $p \sim Uniform(p_{min}, p_{max})$
        \STATE Sample low level policy prior $z_t \sim \pi_{\theta_{high}}(z_t|s_t)$
        \FOR {$t'=t$ \TO $t + p$}
        \STATE Sample action $a_{t'} \sim \pi_{\theta_{low}}(a_{t'}|s_{t'}, z_t)$
        \STATE Sample partner action $a^p_{t'} \sim \pi_{p_n}(a_{t'}|s_{t'})$
        \STATE Obtain new state $s_{t'+ 1}$ and environment reward $r_{t'}$ according to joint action $(a_{t'}, a^p_{t'})$
        \STATE Calculate influence reward $r^{inf}_{t'}$ according to (2)
        \STATE Set $r^l_{t'} \gets r_{t'}$
        \ENDFOR
        \STATE Set $t \gets t +p$
        \STATE Calculate high-level reward according to (3)
        \ENDFOR
    \end{algorithmic}
    \textbf{Output}:\\ high level trajectory: $(s_0,z_0,r^{h}_{0},s_{p_1},z_{p_1},r^{h}_{p_1},..., s_{p_m})$, \\low-level trajectory: $(s_0,a_0,r^{l}_{0},s_1,a_1,r^{l}_{0},..., s_{T})$
\end{algorithm}

\subsection{Training Objective}

Following the same procedure as in HiPPO, we then train both the high-level policies and low-level policies on their respective trajectories with the clipped PPO objective \cite{schulman2017proximal}:

 \begin{equation}
    \begin{aligned}
        L^{CLIP}(\theta) = \hat{E}_t [min\{r_t(\theta)\hat{A}_t, clip(r_t(\theta), 1- \epsilon, 1+\epsilon)\hat{A}_t\}]
    \end{aligned}
\end{equation}

\noindent where $\hat{A}_t$ is the estimated advantage at time $t$ computed with Generalized Advantage Estimation (GAE) \cite{schulman2015high}, $r_t(\theta)$ is the ratio of probabilities under the new and old policies and $\epsilon$ is the clipping parameter. The entire agent is trained end-to-end by summing up the objectives of the high and low-level policies.

\section{Experiments}

\textbf{Environment}. We focus the evaluation of our method on the open-sourced\footnote{\href{https://github.com/HumanCompatibleAI/overcooked\_ai}{https://github.com/HumanCompatibleAI/overcooked\_ai}} two-player Overcooked environment by Carroll et. al. \shortcite{carroll2019utility} based on 
 the collaborative game \textit{Overcooked} \cite{Overcooked}. 

In the game. two agents are tasked to deliver bowls of onion soup through a series of intermediary steps: picking up onions and putting them in the cooking pot for a total of three times, waiting for the soup to finish cooking, picking up a bowl and collecting the soup and finally delivering the soup to the counter. Each successful delivery grants both agents with a +20 reward. The objective of both agents is to maximize the reward in $T=400$ time steps.

We evaluate HiPT on the same set 5 of layouts as \cite{strouse2021collaborating,carroll2019utility}, which are Cramped Room, Asymmetric Advantages, Coordination Ring, Forced Coordination, and Counter Circuit. Visualization of layouts is shown in  Figure \ref{fig:layout}

 
 \medskip

\noindent \textbf{Training and Implementation Details}. We train HiPT on all five layouts for $10^9$ environment steps. We set the upper and lower bound of the execution length of the low-level policy, $(p_{min},p_{max})$ to $[20,40]$ steps. For the high-level policy reward, we set the influence reward coefficient,$\gamma$ from (3) to $1000$ and linearly anneal it to $1$ over the entire training process while the environment reward coefficient $\alpha$ is set to $1$. Further layout-specific implementation details can be found in the supplementary material.

The network architecture of our agent is outlined in Figure \ref{fig:overview}. It consists of 3 initial convolutional layers followed by 2 fully connected layers followed by a recurrent LSTM layer \cite{hochreiter1997long}. The network then splits into the high and low-level policy, both represented by fully connected layers. The prior, $z$ from the high-level policy is represented as a one-hot vector which is then concatenated to the input of the low-level policy. For every layout except Forced Coordination and Counter Circuit, we set the number of sub-policies to learn to $4$. For Forced Coordination and Counter Circuit, we set it to $5$ and $6$ respectively, corresponding to the number of distinct play styles in each population as reflected in Figure \ref{fig:spheatmap}.

\medskip

\noindent \textbf{Comparison Baselines}
We compare HiPT with two separate baseline methods:

\begin{itemize}
    
\item $\boldsymbol{PPO_{BC}}$ \cite{carroll2019utility} is an agent trained using the standard PPO algorithm but with a behavioral cloned Human Model, abbreviated as BC as its partner during training. The BC human model in turn is trained using human-human trajectory data collected by Carroll et.al. We train a $PPO_{BC}$ for each of the 5 different Overcooked layouts.

\item \textbf{Fictitious Co-Play (FCP)} \cite{strouse2021collaborating}. For our comparison, we train the FCP baseline on the same partner population as HiPT to ensure a fair comparison, i.e. we reduce the number of partners from 32 in the original work to 16 but encourage also more diversity in terms of different play-style in the partner population by adding the JSD loss.
\end{itemize}


\subsection{Evaluation with Novel Trained Agents}

We first evaluate HiPT and the baseline comparisons when partnering with two different types of novel trained partner agents:

\medskip

\noindent \textbf{Diverse Self Play Partners}. We trained a separate population of diverse agent partners  of size $16$ with  Self-Play following the same procedure outlined in Section 4.1 to act as an evaluation set of partners. To simulate partners of varying skill levels, we evaluate each of the methods on three versions of each SP partner: Fully trained, a past checkpoint with an intermediate return, and a randomly initialized version, following the evaluation scheme in \cite{strouse2021collaborating}. 

We average the return across all three skill levels and across both starting positions, (blue and green hat) on all five layouts and present the results in Table \ref{tab:selfplay}.

\medskip

\begin{table}
    \centering
    \begin{tabular}{c|ccc}
        \toprule
        Layout  & $PPO_{BC}$  & FCP  & HiPT\\
        \midrule
        Cramped Room     & $106.25$   & $172.5$ & $\boldsymbol{174.71}$        \\
        Asymmetric Advantages   &   $125.83$   & 256.67  & $\boldsymbol{266.88}$      \\
        Coordination Ring  & $28.34$   & $116.58$  & $\boldsymbol{128.54}$  \\
        Forced Coordination & $31.6$   & $69.38$  & $\boldsymbol{75.63}$    \\
        Counter Circuit    & $2.71$   & $65.63$  & $\boldsymbol{106.46}$         \\
        \bottomrule
    \end{tabular}
    \caption{\textbf{Collaboration reward with Diverse Self-Play agents}. Average reward of each type of agent when partnered with a population of Self-play agents. Rewards are over games of length $T=400$. Our method outperforms both FCP and $PPO_{BC}$ at every layout}
    \label{tab:selfplay}
\end{table}


\noindent \textbf{Human Proxy Partners} ($\boldsymbol{H_{proxy}}$) We follow the procedure outlined in \cite{carroll2019utility} to train a behaviorally cloned human model using the open-sourced human-human trajectory data\footnote{\href{https://github.com/HumanCompatibleAI/human\_aware\_rl/}{https://github.com/HumanCompatibleAI/human\_aware\_rl/}}. Crucially the set of human trajectories used to train $H_{proxy}$ and the data used to train the $BC$ model are disjoint so as to ensure that $PPO_{BC}$ would not have an unfair advantage of having access to the evaluation model during training. For $H_{proxy}$, we similarly average the average return across both starting positions for all 5 layouts and compile them in Table \ref{tab:humanproxy}. 

\begin{table}
    \centering
    \begin{tabular}{c|ccc}
        \toprule
        Layout  & $PPO_{BC}$  & FCP  & HiPT\\
        \midrule
        Cramped Room     & $113$   & $148$ & $\boldsymbol{157}$        \\
        Asymmetric Advantages   &   $200$   & 262  & $\boldsymbol{267}$      \\
        Coordination Ring  & $112$   & $121$  & $\boldsymbol{149}$  \\
        Forced Coordination & $48$   & $44$  & $\boldsymbol{51}$    \\
        Counter Circuit    & $32$   & $39$  & $\boldsymbol{48}$         \\
        \bottomrule
    \end{tabular}
    \caption{\textbf{Collaboration reward with Human Proxy agent}. The average reward of each type of agent when partnered with $H_{proxy}$. Rewards are over games of length $T=400$. Our method outperforms both FCP and $PPO_{BC}$ at every layout.}
    \label{tab:humanproxy}
\end{table}

Our results show that HiPT outperforms both the modified version of FCP and $PPO_{BC}$ when partnered with both types of trained agent partners.

\begin{table*}[h]
    \centering
    \begin{tabular}{c|cc|cc}
        \toprule
        \multirow{2}{*}{Layout}& \multicolumn{2}{c|}{$H_{proxy}$} & \multicolumn{2}{c}{ diverse SP} \\
          & with Influ rew.  & w/o Influ rew.  & with Influ rew.  & w/o Influ rew. \\
        \midrule
        Cramped Room     & $\boldsymbol{157.00}$   & $155.00$ & $\boldsymbol{174.71}$  &    $174.58$  \\
        Asymmetric Advantages   &   $\boldsymbol{267.00}$   & $238.00$  & $\boldsymbol{266.88}$ &  $255.63$   \\
        Coordination Ring  & $\boldsymbol{149.00}$   & $130.00$  & $\boldsymbol{128.54}$ & $106.25$\\
        Forced Coordination & $\boldsymbol{51.00}$   & $36.00$  & $75.63$  &  $\boldsymbol{80.42}$\\
        Counter Circuit    & $\boldsymbol{48.00}$   & $21.00$  & $\boldsymbol{106.46}$  &    $55.00$   \\
        \bottomrule
    \end{tabular}
    \caption{\textbf{Comparison of Average Reward between HiPT agents trained with or without Influence Reward}. }
    \label{tab:influencereward}
\end{table*}

\subsection{Ablation study on the effect of adding the Influence Reward}

We also conduct an ablation study to study the effect of the addition of the influence reward to the high-level policy in HiPT. For each of the 5 different Overcooked layouts, we trained a modified version of the HiPT, with the only change being the absence of the influence reward to the high-level policy. We then partnered this reduced version HiPT with both the evaluation SP partners and $H_{proxy}$ partners and compare the average return with the full HiPT agent with the influence reward. We present the results of the ablation study in Table \ref{tab:influencereward}.

Our results show that with the exception of partnering with diverse SP partners in the Forced Coordination layout, the addition of influence reward on the high-level results in consistently better average returns across all layouts when partnered with all types of novel partners.

\begin{figure*}[h]
\vspace{-10pt}
  \includegraphics[width=1\linewidth]{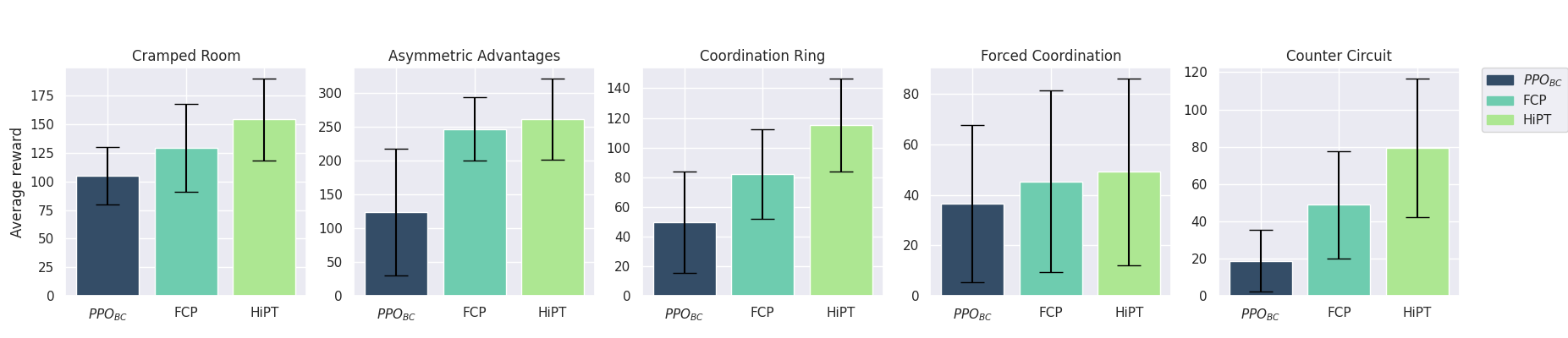}
  \caption{\textbf{Performance of different AI models partnering with human player per layout}}
  \label{fig:human_performance}
   \vspace{-10pt}
\end{figure*}

\begin{figure*}[!htb]
  \centering
  \subfigure[All Participants]{\includegraphics[width=0.30\linewidth]{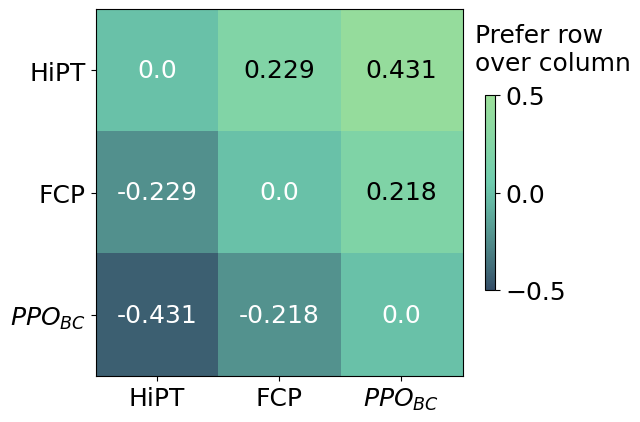} \label{fig:human_preference_a}}
  \subfigure[Very Experienced Participants]{\includegraphics[width=0.30\linewidth]{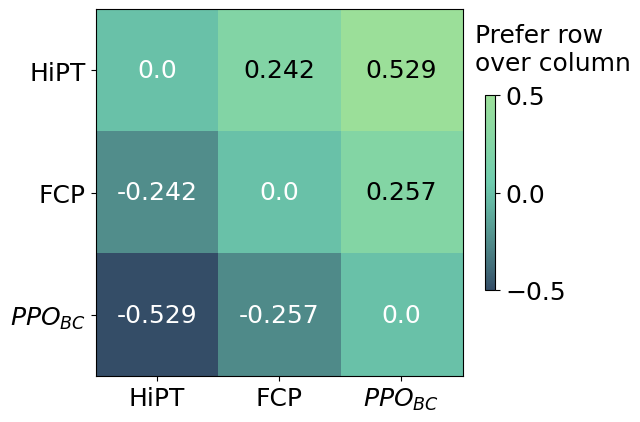}\label{fig:human_preference_b}}
  \subfigure[Moderately Experienced Participants]{\includegraphics[width=0.30\linewidth]{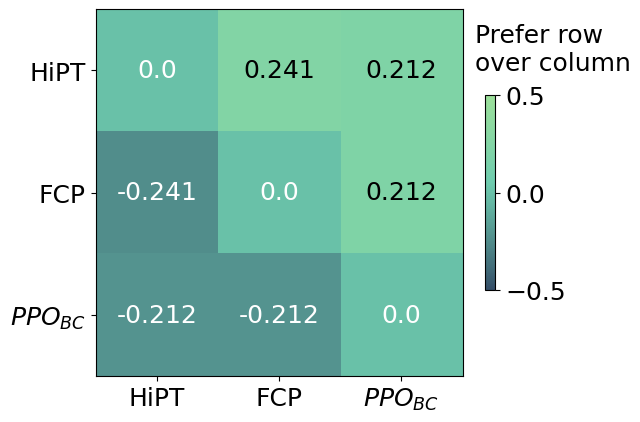}\label{fig:human_preference_c}}

  \caption{\textbf{Participant preference matrix.}The values indicate how much the participants prefer the column agents versus row agents.}
  \label{fig:human_preference}
  \vspace{-10pt}
\end{figure*}

\subsection{Evaluation with Real Human Partners}
We then conduct a user study to evaluate the performance of HiPT as compared other baseline comparisons when partnering with real human partners in Overcooked. We recruit players from Amazon Mechanical Turk\footnote{\href{https://www.mturk.com/}{https://www.mturk.com/}} to participate in our study (N=$72$ ;  $9.7\%$ 
female, $52.8\%$ male, $9.7\%$ others; median age $25-35$ years old). Similar to \cite{strouse2021collaborating}, we used a within-participant design: Each player is invited to play $10$ rounds of games. And, each round consists of $2$ episodes where the layout is kept the same but the AI partners will be different. At the end of every round, the human player will be required to provide their subjective preference on the $2$  agent partners they partnered with.

Upon starting the study, each participant first reads the game instruction and plays 3 tutorial games with an independent AI partner outside of the comparison baselines. Each episode lasts $T=400$ steps (one minute). Then, each participant would play the $10$ pairs of games and report their preference between the $2$ comparing agents on a binary scale (either $-1$ or $+1$ for every preference decision made). The identity of the AI partner is masked so the participant could only give their subjective preferences over the agents based on the agent's partnering capabilities. After completing all the games, the participants are requested to complete a demographic survey to indicate their gender, age group, and to provide a self-evaluation of their experience with video games  as an estimation of the participant's skill level.  

We then perform a repeated measures ANOVA test to compare the effect of AI partners on the reward. There was a statistically significant difference in reward across the players $F(2,142)=138.8133$, $p<0.001$.

\noindent \textbf{Finding One: HiPT model performs best with a human player, achieving the highest averaged reward across all layouts.} We conduct a post-hoc Tukey's Honestly Significant Difference (HSD) test for pair-wise comparison of HiPT against FCP agent and $PPO_{BC}$ respectively, we found that HiPT when paired with human players resulted in higher average rewards across all layouts. HiPT’s mean reward is significantly higher than that of FCP with a mean difference of $19.01$ ($p<0.001$). HiPT’s mean reward is significantly higher than that of $PPO_{BC}$ with a mean difference of $64.49$ ($p<0.001$). We show the results by layout in Figure \ref{fig:human_performance}. In particular, we see a significant improvement for HiPT in the coordination ring and counter circuit layouts when compared with FCP. This resembles the results when validating with a Self-play agent and a Human proxy agent as shown in Table \ref{tab:selfplay} and Table \ref{tab:humanproxy}.  This validates our proposed idea that a hierarchical agent structure increases the robustness of the agent's collaborative capability.

\noindent \textbf{Finding Two: Participants prefer HiPT agent as a partner over all other agents in the preference survey.} Outside of the collaborative capability in terms of total reward, we also study the subjective partner preference of the human participants. Overall, HiPT is the most preferred AI partner in our human studies as shown in Figure \ref{fig:human_preference_a}. The scale shows the averaged preference between $2$ agents on a binary scale. We see a strong correlation between the averaged reward and the participants' preference for an agent. The participants prefer AI partners which could yield a higher reward. We further break down participants' preferences by their self-reported video game experience in Figure \ref{fig:human_preference_b} and Figure \ref{fig:human_preference_c} for very experienced and moderately experienced. Very experienced participants make up the highest portion of the overall number of participants $64.6\%$ while we only receive very few ($4$) participants that self-report to have low experience hence we omit the results for that class of participants. In both classes of participants, HiPT remains the preferred partner. We also break down the average reward with human partners by their self-evaluated experience level in the supplementary material.

\section{Conclusion}

In this work, we introduced Hierarchical Population Training (HiPT), a Hierarchical Reinforcement Learning method for Human-AI Collaboration. HiPT builds upon previous population-based training methods for cooperative agents by extending the diversity of population partners to include partners of diverse play styles and skill levels. In addition, a bi-level hierarchical architecture in HiPT to allow it to learn multiple best-response sub-policies to the partner population in the low-level policy.

We show training HiPT with a diverse partner population outperforms other similar methods when partnering with both diverse trained agent partners and real human partners. We also show that human partners prefer HiPT over other types of agents in our human study. 

For potential future directions, we would like to how to leverage hierarchical agents in collaborative settings to provide a better explanation of the behaviors of both the agents but also of the partner. As the latent sub-policy $z$ encodes the best response policy to a specific play style. We theorize that it should be possible to use such information to produce more explainable agents both on the side of the hierarchical agent and also to classify the play style or behaviors of the partner agent.

\clearpage

\section*{Acknowledgments}

This research was supported by the National Research Foundation, Prime Minister’s Office, Singapore, under its AI Singapore Program (AISG Award No: AISG2-RP-2020-016).

\bibliographystyle{named}
\bibliography{ijcai23}

\clearpage

\appendix
\section{Additional Implementation Details}

In this section, we provide additional implementation details of HiPT.  For all layouts, we set the entropy loss coefficient in the PPO training objective to $0.01$ for both high and low-level policies and linearly decay the value over the course of training. We set the value function coefficient to $0.5$ and the clipping coefficient to $0.05$ for all layouts.

\subsection{Shaped Rewards}

In addition to the +20 reward the agents receive when a soup is delivered to the counter, we also add two more categories of shaped rewards to HiPT agents over the course of training. The first category is the \textbf{Ingredient Pickup/Dropoff Reward} where agents are awarded a +3 reward for every onion picked up and onion dropped into a cooking pot. The second category is the \textbf{Delivery Penalty Reward} where an agent is given a -20 reward penalty if the other agent delivers a soup to the counter. We add the shaped rewards to the training process of every layout except for Forced Coordination as the two different agents are already limited in the tasks they can perform due to the layout. We found empirically that adding these shaped rewards encourages the agents to learn all the different sub-tasks in the Overcooked environment.

\subsection{Layout Specific training  Parameters}

\begin{table}[h]
    \centering
    \begin{tabular}{c|cc}
        \toprule
          & Learning Rate\  &  Learning Rate Decay  \\
        \midrule
        Cramped Room     & $10^{-3}$   & $3$  \\
        Asymmetric Advantages   &   $1^{-3}$   & $3$   \\
        Coordination Ring  & $6^{-4}$   & $1.5$  \\
        Forced Coordination & $8^{-4}$   & $2$  \\
        Counter Circuit    & $8^{-4}$   & $3$   \\
        \bottomrule
    \end{tabular}
    \caption{Layout Specific Training parameters for HiPT. Learning Rate denotes the initial starting learning rate while learning rate decay denotes the ratio in which the learning rate linearly decays to. For example, a layout with learning rate $m$ and decay value $n$ has the learning rate start at $m$ and linearly decay to $\frac{m}{n}$.}
    \label{tab:influencereward}
\end{table}


\section{Additional Results for Real Human Partners by Skill Level}

\begin{figure*}[th]
  \centering
   \subfigure[Very Experienced players] {\includegraphics[width=1\linewidth]
    {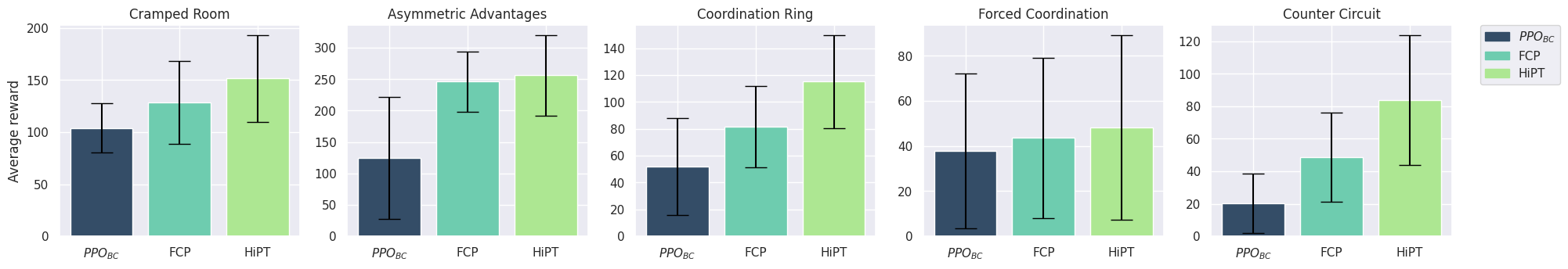}}
   \subfigure[Moderately Experienced Players] {\includegraphics[width=1\linewidth]{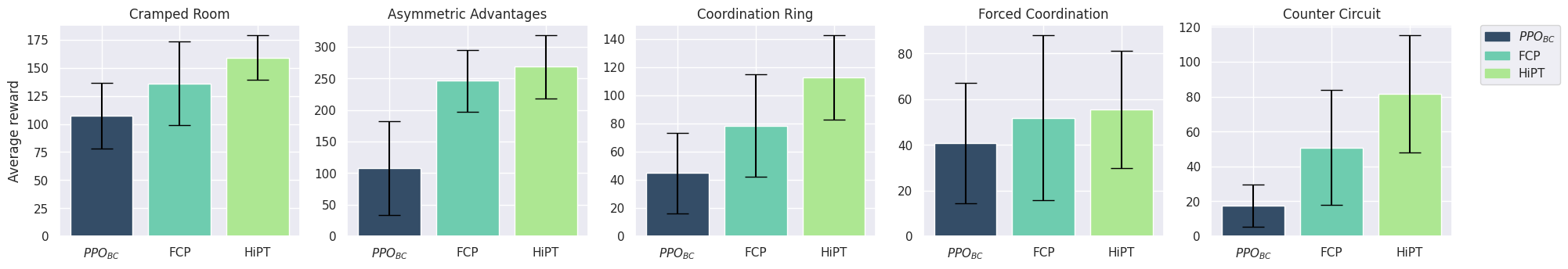}}
   \subfigure[Low Experience Players]  
    {\includegraphics[width=1\linewidth]
    {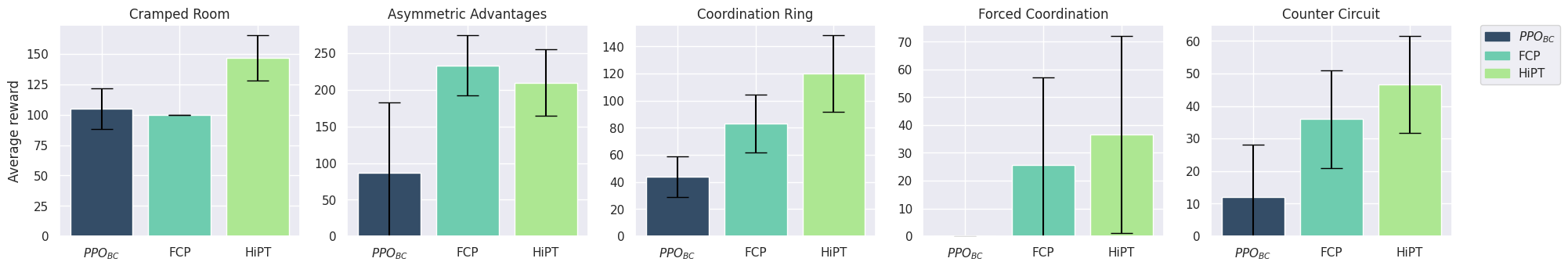}}
  \caption{\textbf{Performance of different AI models partnering with human players per layout at different (self-evaluated) partner experiences with video games}}
  \label{fig:human_performance_by_skill}
\end{figure*}

Here we provide additional results of our experiments with real human partners broken down by skill level (see Figure \ref{fig:human_performance_by_skill}).

\end{document}